\documentclass[12pt]{article}
\usepackage{sbc-template}
\usepackage{graphicx,url}
\usepackage[utf8]{inputenc}

\newcommand{\M}[1]{\mbox {\textit{#1}}}

\title{Prompt-based mental health screening from social media text}

\author{Wesley Ramos dos Santos\inst{1} \and   \\
	Ivandr\'e Paraboni\inst{1} }
\address{University of S\~ao Paulo (EACH-USP) \\ 
	Av Arlindo Bettio 1000, S\~ao Paulo, Brazil  \\
	\email{\{wesley.ramos.santos,ivandre\}@usp.br}}

\begin{document} 

\maketitle

\begin{abstract}
This article presents a method for prompt-based mental health screening from a large and noisy dataset of social media text. Our method uses GPT 3.5. prompting to distinguish publications that may be more relevant to the task, and then uses a straightforward bag-of-words text classifier to predict actual user labels. Results are found to be on pair with a BERT mixture of experts classifier, and incurring only a fraction of its training costs.
\end{abstract}

\section{Introduction}
\label{sec.intro}

Textual representation models used in NLP applications have changed dramatically in recent years, moving on from simple token counts (e.g., bag-of-words)  \cite{jnina} to transformer-based   and large language models (LLMs) such as GPT\footnote{\url{https://platform.openai.com/docs/models}}  and Bloom\footnote{\url{https://huggingface.co/bigscience/bloom}}, among many others. In particular, LLMs have successfully circumvented the need for labelled training data entirely, in so-called  prompt-based methods that are now mainstream in the field, and have been shown to  obtain higher accuracy in a wide range of tasks. 

Despite its potential to reduce training costs, however, LLM prompting still requires careful consideration in applications involving  large or noisy data, as it is often the case of social media. In particular, we notice that social media data may be available as a long history of publications, often stretching over thousands of posts, and yet many or most  may be unrelated to the underlying task. This may be the case, for instance, when screening for user-level information such as mental health statuses (e.g., related to depression), a task that may challenge supervised and prompt-based methods alike.

Based on these observations, this work investigates the use of LLM prompting as an aid to mental health screening from  social media text. Taking the issue of depression detection from Brazilian Twitter timelines as a case  study, we propose a method that combines GPT prompting and standard bag-of-words classification for fast, computationally inexpensive results, and which is  evaluated against SOTA results obtained by BERT mixture of experts  \cite{setembrobr-moe}.

\section{Related work}
\label{sec.related}

Depression detection may be seen as an instance of author profiling \cite{ca-rangel2020,jiis-arthur,brmoral,bracis-pavan} from social media text. Recent studies in the field are summarised in Table \ref{tab.related}, categorised by text genre (Reddit, Twitter), text model  (b=bag of words, BERT, e=embeddings, s=sentiment), and methods.

\begin{table}[ht]
\caption{Depression prediction from text data.}
\centering
\label{tab.related}
\begin{tabular}{ l l l  l}
\hline
Study     	                & Genre        & Text model  & Methods \\
\hline
\cite{smhd} 		  	    & reddit        & b,e 		       & FastText \\
\cite{aragon2019} 	 	 	& reddit 	    & s 			   & SVM \\
\cite{ss3} 			 	 	& reddit 	    & b 			   & SS3 \\
\cite{sensemood} 	 	 	& twitter 	    & e 			   & CNN\\
\cite{comorbidity} 	 	 	& reddit 	    & e 			   & LSTM\\
\cite{souza2021deep}        & reddit        & e                & LSTM+CNN \\
\cite{ansari2022ens}        & reddit,twitter& e,s              & LR+LSTM \\
\cite{setembrobr}           & twitter       & BERT             & Bi-LSTM\\
\cite{setembrobr-moe}       & twitter       & BERT             & mixture of experts\\
\hline
\end{tabular}
\end{table}

We notice a slight predominance of works based on Reddit. This may be explained by the greater ease of access and reuse of this type of data for research purposes, which is far more restricted in the case of  Twitter/X. Reddit publications are taken as the basis for some of the best-known datasets available for the English language, including the SMHD \cite{smhd} and eRisk \cite{erisk2022} corpora.

As for the kinds of textual model under consideration, Table \ref{tab.related} reflects the natural evolution of the NLP and related fields, with an initial prevalence of bag-of-words models and feature engineering, and its gradual replacement by models based on word embeddings and, more recently, transformers. With regard to the computational methods used, a similar trajectory is generally observable, with the use of traditional (e.g., linear) classifiers  being gradually replaced by sequence classification methods based on deep learning, including the more recent  use of transformer-based architectures. In both cases - text representation and methods - we notice also that better results are usually accompanied by an increase in computational costs. 

Finally, we notice that, with the exception of the work based on the \M{SetembroBR} \cite{setembrobr,depress-lrec} corpus to be used in the present work, all of the above studies are devoted to the English language and, to the best of our knowledge, none make use of prompt-based methods to query a LLM  for depression directly. 

\section{Method}
\label{sec.method}

As in \cite{smhd,erisk2022} and others, our approach to depression detection  relies on a dataset of social media (in our case,  Twitter/X) texts that have been published by either  individuals who self-reported a depression diagnosis, or by a random (control) group. Thus, the  task at hand represents a binary classification problem intended  to distinguish  individuals who will most likely receive a depression diagnosis (called Diagnosed class) in the future from the general population (called Control class), and not to distinguish depressed from non-depressed individuals per se\footnote{In fact, the data does not convey any guaranteed `non-depressed' individuals \cite{setembrobr}.}. %


\subsection{Data}
\label{sec.data}

We  use of the depression portion of the SetembroBR corpus \cite{setembrobr}, a collection of Twitter timelines  published  by Diagnosed and Control individuals in which only the data prior to the diagnosis date is kept. As in other language resources of this kind, the Control (i.e., random) subset is designed so as to be seven times larger than the Diagnosed group, making a heavily imbalanced classification task. Table \ref{tab.stats.corpus} presents data descriptive statistics.

\begin{table}[ht]
\caption{\mbox{SetembroBR} depression corpus descriptive statistics.}
\centering
\label{tab.stats.corpus}
\begin{tabular}{ l    c c    c }
\hline
Statistics               & Diagnosed    & Control  & Overall\\
\hline
Users (timelines)        & 1,684        & 11,788   & 13,472 \\
{Words} (million)	     & 29.32		& 201.94   & 231,26 \\
{Publications} (million) & 2.43		    & 16.99    & 19,42 \\
\hline
\end{tabular}
\end{table}

In the present work, we follow the standard train/test split provided by the corpus, as described in \cite{setembrobr}.

\subsection{Approach}
\label{sec.approach}

Screening for depression from social media gives rise to the questions of how to handle a large number of publications, most of which unlikely to be relevant to the task. To this end,  we envisaged a prompt-based approach called \M{Prompt.Bow}  that relies on GPT 3.5 to enrich an otherwise  standard text classifier. These two steps - prompting and classification - are described individually as follows. 

\begin{figure}[ht]
\centering
\includegraphics[width=1\textwidth]{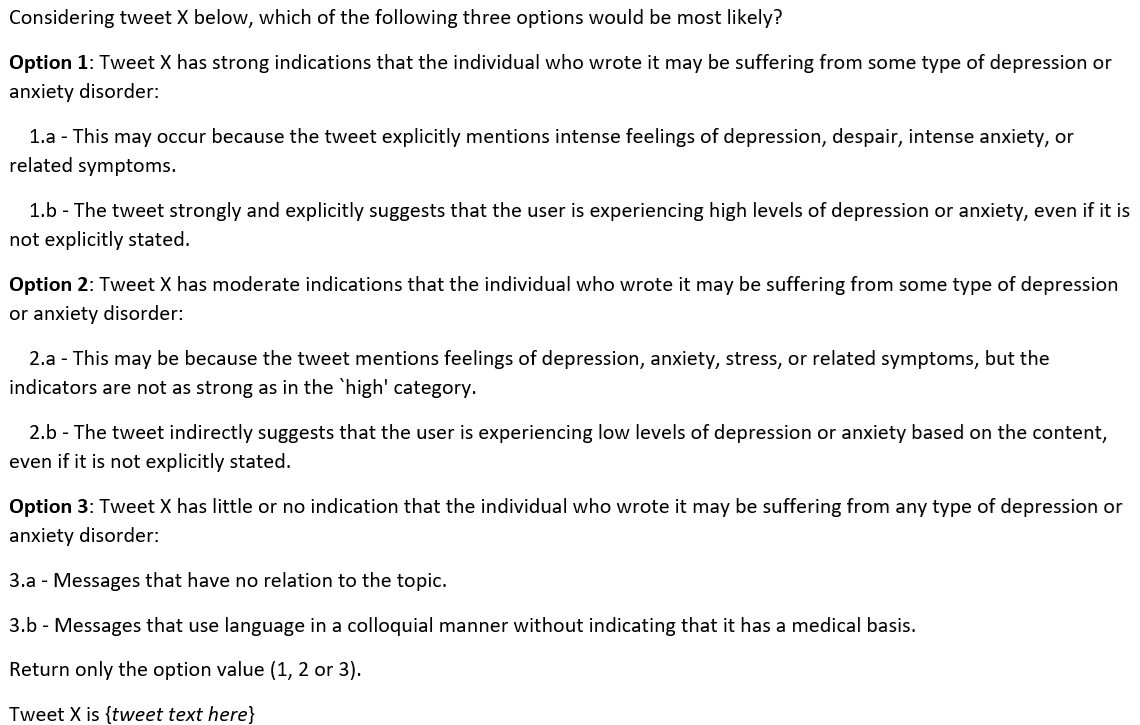}
\caption{Prompt instruction to GPT.}
\label{fig.prompt}
\end{figure}

First as a means to identify messages that are potentially related to mental health, we use  GPT 3.5. prompting  to assess a random sample of 30,000 tweets. The (English-translated) prompt in question, adapted from the clinical description of depression in \cite{dm5}, is shown in Figure \ref{fig.prompt}.

By submitting this prompt to each of the 30,000 sample tweets, the data was categorised as having high (1), medium (2) or low (3) relevance to mental health. These GPT-labelled data was then taken as an input to train a T5 classifier \cite{t5} to label the entire 19.42-million tweets in the corpus. Table shows the label distribution across the training portion of the data.

\small{
\begin{table}[ht]
\caption{Training data distribution according to relevance for depression. Tweets and tokens are shown in thousand units.}
\centering
\setlength{\tabcolsep}{5pt}
\label{tab.gptdist}
\begin{tabular}{l | rcrc | rcrc}
\hline
\multicolumn{1}{c|}{}&\multicolumn{4}{c|}{Diagnosed class}&\multicolumn{4}{c}{Control class}\\
Relevance   & Tweets  & \%     & Tokens    & \%       & Tweets  & \%      & Tokens      & \%\\
\hline
high        & 39    & 2.0\%  & 674   & 5.4\%    &    153 & 1.1\%  &  2,372 & 3.0\%\\
medium      & 180   & 9.3\%  & 3,663 & 29.4\%   &  1,031 & 7.6\%  & 20,566 & 25.7\%\\
low         & 1,727 & 88.7\% & 8,122 & 65.2\%   & 12,439 & 91.3\% & 56,971 & 71.3\%\\
\hline\end{tabular}
\end{table}
}
\normalsize


As expected, most publications (65.2\% in the Diagnosed class, and 71.3\% in the Control class) are deemed of low relevance for mental health prediction according to the prompted model. On the other hand, highly relevant publications are relatively rare (5.4\% in the Diagnosed class, and 3.0\% in the Control class). 

More importantly, however, when prompting an LLM for mental health we are largely focusing on semantics, that is, on symptoms and other well-known clinical signs of depression. For instance, our current prompt allows the model to pinpoint a wide range of publications that may suggest, e.g.,  eating disorders or negative language use, both of which known to be related to depression, cf. \cite{dm5}, but  this is not to say that other factors can or should be overlooked. 

In particular, we notice that LLM prompting  may not explicitly account for more fine-grained linguistic indicators of depression such as the use of first person pronouns \cite{dep-indicators}, absolute terms \cite{absolute} and  other lexical factors (e.g., denoting emotion as in hate speech, cf. \cite{odio}). These indicators, which may be present in any publication of low or high relevance to depression alike, are also important predictors of mental health statuses, and no data point should in principle be discarded solely based on the LLM output.

As a means to keep the full train data available to the classifier whilst distinguishing between more and less relevant messages (which clearly show different distributions across classes in Table \ref{tab.gptdist}), \M{Prompt.Bow}  uses the category labels provided by the LLM  to split the training data into high, medium, and low relevance subsets, from which we  create three individual bag-of-words vectors (to be concatenated  as discussed below). In other words, the training texts are split into three (low/medium/high relevance) categories according to the previous GPT prompt method, and we build an individual BoW model from each of these three subsets. 

The resulting vectors  are further reduced using univariate feature selection using F1 as a score function. The final `high' and `medium' vectors were reduced to $k=$ 6,000 features each, and the final `low' vector was reduced to $k=$ 3,000 features, all of which concatenated as a single 15,000-word vector. 

In addition to distinguishing between high-, medium- and low-relevance messages in this way, we use the information provided by the LLM also to help capture message order, the underlying assumption being that certain patterns  (e.g., a series of consecutive `highly relevant' messages) may be indicative of depression. To this end, we created a bigram model of high/medium/low labels only, that is, representing sequences of `high', `medium' and `low' labels only (and not text). This was further reduced to its $k=$ 40,000 most relevant (bigram) features by performing F1 univariate feature selection, and then appended to the previous 15,000-word  vector. 

The resulting vector - a combination of three text models with different degrees of relevance to mental health, and a model that captures sequences of relevant messages - is taken as the input to a standard logistic regression classifier. This choice is motivated by the observation that much of the deep (e.g., semantics-driven) language processing had already been performed by the LLM, and that a simple text classifier should suffice for user label prediction.

\section{Evaluation}
\label{sec.results}

Table \ref{tab.results} presents the results of the present \M{Prompt.Bow} model alongside the results reported in \cite{setembrobr-moe} for BERT mixture of experts using BERTabaporu \cite{bertabaporu}, which currently stands as the  SOTA  for the present setting. In both cases, results are based on the standard test portion of the corpus.  
              
\begin{table}[ht]
\caption{Classification results.}
\centering
\label{tab.results}
\begin{tabular}{lccc}
\hline
Model        & Precision        & Recall     & F1    \\
\hline
Prompt.BoW   & 0.64             & 0.72       & 0.66 \\
BERT.MoE     & 0.64             & 0.67       & 0.65 \\ 
\hline\end{tabular}
\end{table}

Table \ref{tab.results} shows that results remain close, with a small advantage for the \M{Prompt.Bow} approach over BERT mixture of experts. However, by relying on an input provided by the pre-trained LLM, we notice that these results were obtained by using a computationally inexpensive classifier model (that is, leaving aside  the costs of pre-training the LLM in the first place), which represents a stark contrasts to computationally-intensive BERT mixture of experts supervision.

\section{Final remarks}
\label{sec.final}

This work presented an experiment in prompt-based mental health screening from a large, noisy corpus of social media publications that relies on LLM prompting to distinguish publications that may be more relevant to the task, and then uses a straightforward  bag-of-words text classifier to predict actual user labels. This was shown to obtain competitive results if compared to a BERT-based classifier architecture that represents the SOTA in the present setting, but incurring only a  fraction of its training costs.

As future work, we intend to refine the present method by using the formal definition of depression in \cite{dm5} as a prompt, and by fully integrating symptoms and linguistic indicators within a neural architecture for both depression and anxiety disorder prediction.   

\pagebreak

\section{Acknowledgements}

The present work has been financed by the S\~ao Paulo Research Foundation (FAPESP grant \#2021/08213-0). The first author has been supported  by the Coordena\c{c}\~ao de Aperfei\c{c}oamento de Pessoal de N\'ivel Superior - Brasil (CAPES) - Finance Code 001 - grant \# 88887.475847/2020-00. This work was carried out at the Center for Artificial Intelligence (C4AI-USP), with support by the S\~ao Paulo Research Foundation (FAPESP grant \#2019/07665-4) and by the IBM Corporation.

\bibliographystyle{sbc}
\bibliography{refs}

\end{document}